\documentstyle[aclap]{article}

\input{psfig}

\newcommand{\ignore}[1]{}

\title{\vspace{-0.5in}  Semi-Automatic Acquisition of \\ Domain-Specific
			Translation Lexicons}
\author{Philip Resnik \\ 
       Dept. of Linguistics and UMIACS \hspace{0.2in} \\ 
       University of Maryland \\
       College Park, MD 20742 USA \\
       {\tt resnik@umiacs.umd.edu}
  \And  
       I. Dan Melamed \\
       \hspace{0.2in} Dept. of Computer and Information Science \\
       University of Pennsylvania \\
       Philadelphia, PA 19104 USA \\
       {\tt melamed@unagi.cis.upenn.edu}
}

\begin{document}
\bibliographystyle{acl}
\maketitle
\vspace{-0.5in}
\begin{abstract}
We investigate the utility of an algorithm for translation lexicon
acquisition (SABLE), used previously on a very large corpus to acquire
general translation lexicons, when that algorithm is applied to a much
smaller corpus to produce candidates for domain-specific translation
lexicons.
\end{abstract}

\section{Introduction}
\label{sec:intro}

Reliable translation lexicons are useful in many applications, such as
cross-language text retrieval.  Although general purpose machine
readable bilingual dictionaries are sometimes available, and although
some methods for acquiring translation lexicons automatically from
large corpora have been proposed, less attention has been paid to the
problem of acquiring bilingual terminology specific to a domain,
especially given domain-specific parallel corpora of only limited
size.

In this paper, we investigate the utility of an algorithm for
translation lexicon acquisition \cite{mel97}, used previously on a
very large corpus to acquire general translation lexicons, when that
algorithm is applied to a much smaller corpus to produce candidates
for domain-specific translation lexicons.  The goal is to produce
material suitable for postprocessing in a lexicon acquisition process
like the following:
\begin{enumerate}
\item Run the automatic lexicon acquisition algorithm on a
      domain-specific parallel corpus.
\item Automatically filter out ``general usage'' entries
      that already appear in a machine readable dictionary (MRD)
      or other general usage lexical resources.
\item Manually filter out incorrect or irrelevant entries from the 
      remaining list.
\end{enumerate}
Our aim, therefore, is to achieve sufficient recall and precision to
make this process --- in particular the time and manual effort required in
Step~3 --- a viable alternative to manual creation of translation lexicons
without automated assistance.

The literature on cross-lingual text retrieval (CLTR) includes work
that is closely related to this research, in that recent approaches
emphasize the use of dictionary- and corpus-based techniques for
translating queries from a source language into the language of the
document collection \cite{Oard97d}.  Davis and Dunning
\shortcite{davis1995:trec4}, for example, generate target-language
queries using a corpus-based technique that is similar in several
respects to the work described here.  However, the approach does not
attempt to distinguish domain-specific from general usage term pairs,
and it involves no manual intervention.  The work reported here,
focusing on semi-automating the process of acquiring translation
lexicons specific to a domain, can be viewed as providing bilingual
{\em dictionary\/} entries for CLTR methods like that used by Davis in
later work \cite{davis1996:trec5}, in which dictionary-based
generation of an ambiguous target language query is followed by
corpus-based disambiguation of that query.

Turning to the literature on bilingual terminology identification {\em
per se}, although monolingual terminology extraction is a problem that
has been previously explored, often with respect to identifying
relevant multi-word terms
(e.g. \cite{daille:balancing_act,smadja1993}), less prior work exists
for bilingual acquisition of domain-specific translations.  {\em
Termight\/} \cite{Dagan/Church:ANLP:94} is one method for analyzing
parallel corpora to discover translations in technical terminology;
Dagan and Church report accuracy of 40\% given an English/German
technical manual, and observe that even this relatively low accuracy
permits the successful application of the system in a translation
bureau, when used in conjunction with an appropriate user interface.

The {\em Champollion\/} system \cite{smadja1996} moves toward higher
accuracy (around 73\%) and considerably greater flexibility in the
handling of multi-word translations, though the algorithm has been
applied primarily to very large corpora such as the Hansards (3-9
million words; Smadja et al. observe that the method has difficulty
handling low-frequency cases), and no attempt is made to distinguish
corpus-dependent translations from general ones.

Daille et al. \shortcite{daille1994} report on a study in which a
small (200,000 word) corpus was used as the basis for extracting
bilingual terminology, using a combination of syntactic patterns for
identifying simple two-word terms monolingually, and a statistical
measure for selecting related terms across languages.  Using a
manually constructed reference list, they report 70\%
precision.

The SABLE system \cite{mel96b} makes no attempt to handle
collocations, but for single-word to single-word translations it
offers a very accurate method for acquiring high quality translation
lexicons from very large parallel corpora: Melamed reports 90+\%
precision at 90+\% recall, when evaluated on sets of Hansards data of
6-7 million words.  Previous work with SABLE does not attempt to
address the question of domain-specific vs. general translations.

This paper applies the SABLE system to a much smaller (approximately
400,000 word) corpus in a technical domain, and assesses its potential
contribution to the semi-automatic acquisition process outlined above,
very much in the spirit of Dagan and Church
\shortcite{Dagan/Church:ANLP:94} and Daille et
al. \shortcite{daille1994}, but beginning with a higher accuracy
starting point and focusing on mono-word terms.  In the remainder of
the paper we briefly outline translation lexicon acquisition in the
SABLE system, describe its application to a corpus of technical
documentation, and provide a quantitative assessment of its
performance.

\section{SABLE}
\label{sec:SABLE}

SABLE (Scalable Architecture for Bilingual LExicography) is a turn-key
system for producing clean broad-coverage translation lexicons from
raw, unaligned parallel texts (bitexts).  Its design is modular and
minimizes the need for language-specific components, with no
dependence on genre or word order similarity, nor sentence boundaries
or other ``anchors'' in the input.

SABLE was designed with the following features in mind:
\begin{itemize}

\item {\em Independence from linguistic resources}: SABLE does not
rely on any language-specific resources other than tokenizers and a
heuristic for identifying word pairs that are mutual translations,
though users can easily reconfigure the system to take advantage of
such resources as language-specific stemmers, part-of-speech taggers,
and stop lists when they are available.

\item {\em Black box functionality}: Automatic acquisition of
translation lexicons requires only that the user provide the input
bitexts and identify the two languages involved.

\item {\em Robustness}: The system performs well even in the face of
omissions or inversions in translations.

\item {\em Scalability}: SABLE has been used successfully on input
bitexts larger than 130MB.

\item {\em Portability}: SABLE was initially implemented for
French/English, then ported to Spanish/English and to Korean/English.
The porting process has been standardized and documented
\cite{mel96c}.

\end{itemize}

The following is a brief description of SABLE's main components.  A
more detailed description of the entire system is available in
\cite{mel97}.

\subsection{Mapping Bitext Correspondence}
\label{map}
After both halves of the input bitext(s) have been tokenized, SABLE
invokes the {\em Smooth Injective Map Recognizer (SIMR)} algorithm
\cite{mel96a} and related components to produce a bitext map.  A
bitext map is an injective partial function between the character
positions in the two halves of the bitext.  Each point of
correspondence $(x,y)$ in the bitext map indicates that the word
centered around character position $x$ in the first half of the bitext
is a translation of the word centered around character position $y$ in
the second half.  SIMR produces bitext maps a few points at a time, by
interleaving a point generation phase and a point selection phase.

SIMR is equipped with several ``plug-in'' matching heuristic modules
which are based on cognates \cite{davis1995:eacl,simard,mel95} and/or
``seed'' translation lexicons \cite{chen}.  Correspondence points are
generated using a subset of these matching heuristics; the particular
subset depends on the language pair and the available resources.  The
matching heuristics all work at the word level, which is a happy
medium between larger text units like sentences and smaller text units
like character n-grams.  Algorithms that map bitext correspondence at
the phrase or sentences level are limited in their applicability to
bitexts that have easily recognizable phrase or sentence boundaries,
and Church \shortcite{charalign} reports that such bitexts are far
more rare than one might expect.  Moreover, even when these larger
text units can be found, their size imposes an upper bound on the
resolution of the bitext map.  On the other end of the spectrum,
character-based bitext mapping algorithms
\cite{charalign,davis1995:eacl} are limited to language pairs where
cognates are common; in addition, they may easily be misled by
superficial differences in formatting and page layout and must
sacrifice precision to be computationally tractable.

SIMR filters candidate points of correspondence using a geometric
pattern recognition algorithm.  The recognized patterns may contain
non-monotonic sequences of points of correspondence, to account for
word order differences between languages.  The filtering algorithm can
be efficiently interleaved with the point generation algorithm so that
SIMR runs in linear time and space with respect to the size of the
input bitext.

\subsection{Translation Lexicon Extraction}
\label{tralex}

\begin{figure}%[htb]
\centerline{\psfig{figure=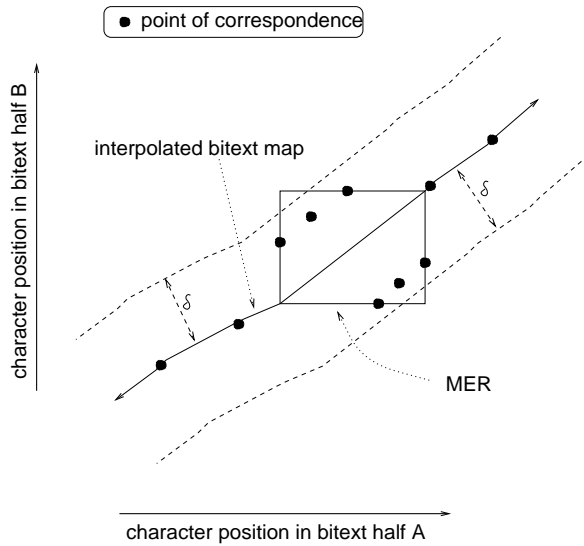,width=3in}}
\caption{{\em Word token pairs whose co-ordinates lie between the dashed
boundaries count as co-occurrences.}}
\label{delta}
\end{figure}
Since bitext maps can represent crossing correspondences, they are
more general than ``alignments'' \cite{mel96a}.  For the same
reason, bitext maps allow a more general definition of token
co-occurrence.  Early efforts at extracting translation lexicons from
bitexts deemed two tokens to co-occur if they occurred in aligned
sentence pairs \cite{wordcorr}.  SABLE counts two tokens as
co-occurring if their point of correspondence lies within a short
distance $\delta$ of the interpolated bitext map in the bitext space,
as illustrated in Figure~\ref{delta}.  To ensure that interpolation is
well-defined, minimal sets of non-monotonic points of correspondence
are replaced by the lower left and upper right corners of their 
minimum enclosing rectangles (MERs).

SABLE uses token co-occurrence statistics to induce an initial
translation lexicon, using the method described in \cite{mel95}.  The
{\em iterative filtering\/} module then alternates between estimating
the most likely translations among word tokens in the bitext and
estimating the most likely translations between word types.  This
re-estimation paradigm was pioneered by Brown et al. \shortcite{ibm}.
However, their models were not designed for human inspection, and
though some have tried, it is not clear how to extract translation
lexicons from their models \cite{wu95}.  In contrast, SABLE
automatically constructs an explicit translation lexicon, the lexicon
consisting of word type pairs that are not filtered out during the
re-estimation cycle.  Neither of the translation lexicon construction
modules pay any attention to word order, so they work equally well for
language pairs with different word order.

\subsection{Thresholding}
Translation lexicon recall can be automatically computed with respect
to the input bitext \cite{mel96b}, so SABLE users have the option of
specifying the recall they desire in the output.  As always, there is
a tradeoff between recall and precision; by default, SABLE will choose
a likelihood threshold that is known to produce reasonably high
precision.

\section{Evaluation in a Technical Domain}
\label{sec:evaluation}

\subsection{Materials Evaluated}
\label{sec:materials}

\begin{figure}%[htb]
\centerline{\psfig{figure=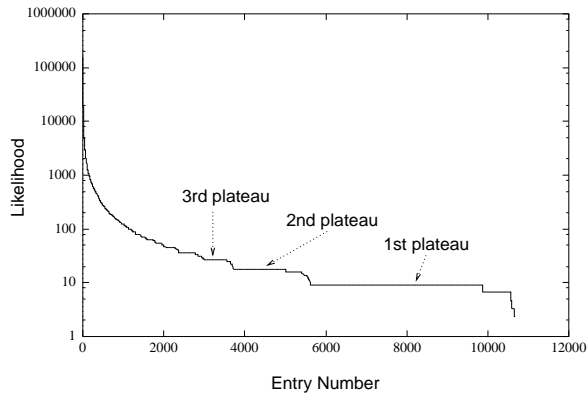,width=3in}}
\caption{{\em Translation lexicon entries proposed by SABLE exhibit
plateaus of likelihood.}}
\label{fig:plateaus}
\end{figure}

The SABLE system was run on a corpus comprising parallel versions of
Sun Microsystems documentation (``Answerbooks'') in French (219,158
words) and English (191,162 words).  As Melamed \shortcite{mel96b}
observes, SABLE's output groups naturally according to ``plateaus'' of
likelihood (see Figure~\ref{fig:plateaus}).  The translation lexicon
obtained by running SABLE on the Answerbooks contained 6663
French-English content-word entries on the 2nd plateau or higher,
including 5464 on the 3rd plateau or higher. Table~\ref{tbl:sample}
shows a sample of 20 entries selected at random from the Answerbook
corpus output on the \mbox{3rd} plateau and higher.  Exact
matches, such as {\em cpio/cpio\/} or {\em clock/clock}, comprised
roughly 18\% of the system's output.
\begin{table}%[htb]
\begin{small}
\begin{center}
\begin{tabular}{|l|l|}\hline
French		& English \\ \hline\hline
constantes		& constants	\\
multi-fen\^etrage	& windows	\\
risque			& may	\\
extensions		& extensions	\\
exemple			& such	\\
rel\^ach\'e		& released	\\
rw-r			& r	\\
re\c{c}us		& received	\\
pr\'ealable		& first	\\
cpio			& cpio	\\
sont			& will	\\
defaults		& defaults	\\
fn			& fn	\\
alphab\'etique		& alphabetically	\\
activ\'ee		& activates	\\
machine			& workstation	\\
mettre			& turns	\\
connect\'es		& connected	\\
bernard			& spanky	\\
superutilisateur	& root	\\
\hline
\end{tabular}
\end{center}
\end{small}
\caption{{\em Random sample of SABLE output on software manuals.}
	  \label{tbl:sample}}
\end{table}

In order to eliminate likely general usage entries from the initial
translation lexicon, we automatically filtered out all entries that
appeared in a French-English machine-readable dictionary (MRD)
\cite{collins1991:french_english}.  4071 entries remained on or above
the 2nd likelihood plateau, including 3135 on the 3rd likelihood
plateau or higher.  

In previous experiments on the Hansard corpus of Canadian
parliamentary proceedings, SABLE had uncovered valid
general usage entries that were not present in the Collins MRD
(e.g. {\em pointill\'es/dotted}).  Since entries obtained from the
Hansard corpus are unlikely to include relevant technical terms, we
decided to test the efficacy of a second filtering step, deleting all
entries that had also been obtained by running SABLE on the Hansards.
On the 2nd plateau or higher, 3030 entries passed both the Collins and
the Hansard filters; 2224 remained on or above the 3rd plateau.

Thus in total, we evaluated four lexicons derived from all
combinations of two independent variables: cutoff (after the 2nd
plateau vs.\ after the 3rd plateau) and Hansards filter (with filter
vs.\ without).  Evaluations were performed on a random sample of 100
entries from each lexicon variation, interleaving the four samples to
obscure any possible regularities.  Thus from the evaluator's
perspective the task appeared to involve a single sample of 400
translation lexicon entries.

\subsection{Evaluation Procedure}
\label{sec:procedure}

Our assessment of the system was designed to reasonably approximate
the post-processing that would be done in order to use this system for
acquisition of translation lexicons in a real-world setting, which
would necessarily involve subjective judgments.  We hired six fluent
speakers of both French and English at the University of Maryland;
they were briefed on the general nature of the task, and given a data
sheet containing the 400 candidate entries (pairs containing one
French word and one English word) and a ``multiple choice'' style
format for the annotations, along with the following instructions.
\begin{quote}
\begin{small}
\begin{enumerate}

\item If the pair clearly cannot be of help in constructing a glossary, 
	circle ``{\bf Invalid}'' and go on to the next pair.

\item If the pair can be of help in constructing a glossary, choose
	{\em one\/} of the following:\footnote{Since part-of-speech
	tagging was used in the version of SABLE that produced the
	candidates in this experiment, entries presented to the
	annotator also included a minimal form of part-of-speech
	information, e.g. distinguishing nouns from verbs.  The
	annotator was informed that these annotations were the
	computer's best attempt to identify the part-of-speech for the
	words; it was suggested that they could be used as a hint as
	to why that word pair had been proposed, if so desired, and
	otherwise ignored.}

\begin{itemize}
  \item[{\bf V}:] The two words are of the ``plain vanilla'' type you might
           find in a bilingual dictionary.

  \item[{\bf P}:] The pair is a case where a word changes its part of speech
           during translation.  For example, ``to have protection'' in
           English is often translated as ``\^etre prot\'eg\'e'' in Canadian
           parliamentary proceedings, so for that domain the pair
           protection/prot\'eg\'e would be marked P.

  \item[{\bf I}:] The pair is a case where a direct translation is incomplete
           because the computer program only looked at single words.
           For example, if French ``imm\'ediatement'' were paired with
           English ``right'', you could select I because the pair is
           almost certainly the computer's best but incomplete attempt
           to be pairing ``imm\'ediatement'' with ``right away''.
\end{itemize}

\item Then choose {\em one\/} or {\em both\/} of the following:

\begin{itemize}

  \item {\bf Specific}.    Leaving aside the relationship between the two words
                  (your choice of P, V, or I), the word pair would be
                  of use in constructing a {\em technical\/} glossary.

  \item {\bf General}.     Leaving aside the relationship between the two words
                  (your choice of P, V, or I), the word pair would be
                  of use in constructing a {\em general usage\/} glossary.
\end{itemize}

    Notice that a word pair could make sense in both.  For example,
    ``corbeille/wastebasket'' makes sense in the computer domain
    (in many popular graphical interfaces there is a wastebasket icon 
    that is used for deleting files), but also in more general usage.
    So in this case you could in fact decide to choose both ``Specific''
    and ``General''.  If you can't choose either ``Specific'' {\em or\/}
    ``General'', chances are that you should reconsider whether or not
    to mark this word pair ``Invalid''.

\item If you're completely at a loss to decide whether or not the word 
	    pair is valid, just put a slash through the number of the example
	    (the number at the beginning of the line) and go on to the next
	    pair.
\end{enumerate}
\end{small}
\end{quote}
Annotators also had the option of working electronically rather than
on hardcopy.

The assessment questionnaire was designed to elicit information
primarily of two kinds.  First, we were concerned with the overall
accuracy of the method; that is, its ability to produce reasonable
candidate entries whether they be general or domain specific.  The
``Invalid'' category captures the system's mistakes on this dimension.
We also explicitly annotated candidates that might be useful in
constructing a translation lexicon, but possibly require further
elaboration.  The {\em V\/} category captures cases that require
minimal or no additional effort, and the {\em P\/} category covers
cases where some additional work might need to be done to accommodate
the part-of-speech divergence, depending on the application.  The {\em
I\/} category captures cases where the correspondence that has been
identified may not apply directly at the single-world level, but
nonetheless does capture potentially useful information.  Daille et
al. \shortcite{daille1994} also note the existence of ``incomplete''
cases in their results, but collapse them together with ``wrong''
pairings.

Second, we were concerned with domain specificity.  Ultimately we
intend to measure this in an objective, quantitative way by comparing
term usage across corpora; however, for this study we relied on human
judgments.

\subsection{Use of Context}

Melamed \shortcite{mel96b} suggests that evaluation of translation
lexicons requires that judges have access to bilingual concordances
showing the contexts in which proposed word pairs appear; however,
out-of-context judgments would be easier to obtain in both
experimental and real-world settings.  In a preliminary evaluation, we
had three annotators (one professional French/English translator and
two graduate students at the University of Pennsylvania) perform a
version of the annotation task just described: they annotated a set of
entries containing the output of an earlier version of the SABLE
system (one that used aligned sub-sentence fragments to define term
co-occurrence; cf. Section~\ref{tralex}).  No bilingual concordances
were made available to them.

Analysis of the system's performance in this pilot study, however, as
well as annotator comments in a post-study questionnaire, confirmed
that context is quite important.  In order to quantify its importance,
we asked one of the pilot annotators to repeat the evaluation on the
same items, this time giving her access to context in the form of the
bilingual concordances for each term pair.  These concordances
contained up to the first ten instances of that pair as used in
context.  For example, given the pair {\em d\'eplacez/drag}, one
instance in that pair's bilingual concordance would be:
\begin{itemize}
\item[] Maintenez SELECT enfonc\'e et {\bf d\'eplacez} le dossier vers
        l' espace de travail .
\item[] Press SELECT and {\bf drag} the folder onto the workspace
        background .
\end{itemize}
The instructions for the in-context evaluation specify that the
annotator should look at the context for every word pair, pointing out
that ``word pairs may be used in unexpected ways in technical text and
words you would not normally expect to be related sometimes turn out
to be related in a technical context.''

Although we have data from only one annotator, Table~\ref{tbl:context}
shows the clear differences between the two results.\footnote{Again,
this sample of data was produced by an older and less accurate version
of SABLE, and therefore the percentages should only be analyzed
relative to each other, not as absolute measures of performance.}
\begin{table*}[htb]
\centering
\begin{tabular}{|c||c|c|c|c||c|c|c|} \hline
& \% & \% & \% & All Valid & Domain-Specific & General Usage & Both \\
& V  & P  & I  & Entries   & Only            & Only & \\
\hline\hline
Out-of-Context & 
  39.5 & 9.25  & 5.5  & 57.75 & 29.75  & 23.5  & 1 \\
In-Context & 
  46.75 & 5  & 13 & 69.5 & 38  & 23.25  & 3.5 \\
\hline
\end{tabular}
\caption{{\em Effect of in-context vs. out-of-context evaluation. All
numbers are in \%. $n$ = 400.} \label{tbl:context}}
\end{table*}
In light of the results of the pilot study, therefore, our six
annotators were given access to bilingual concordances for the entries
they were judging and instructed in their use as just described.

\section{Results}

\subsection{Group Annotations}

A ``group annotation'' was obtained for each candidate translation
lexicon entry based on agreement of at least three of the six
annotators.  ``Tie scores'' or the absence of a 3-of-6 plurality were
treated as the absence of an annotation.  For example, if an entry was
annotated as ``Invalid'' by two annotators, marked as category V and
Specific by two annotators, and marked as category P, Specific, and
General by the other two annotators, then the group annotation would
contain an ``unclassified valid type'' (since four annotators chose a
valid type, but there was no agreement by at least three on the
specific subclasification) and a ``Specific'' annotation (agreed on by
four annotators).  All summary statistics are reported in terms of the
group annotation.

\subsection{Precision} 
\label{precision}

\begin{figure}[htb]
\centerline{\psfig{figure=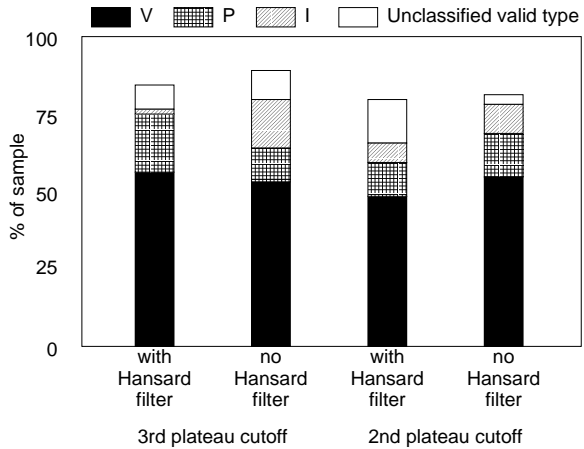,width=3in}}
\caption{{\em Summary of filtered translation lexicon validity statistics.}}
\label{valid}
\end{figure}

SABLE's precision on the Answerbooks bitext is summarized in
Figure~\ref{valid}.\footnote{The exact numbers gladly provided on
request.}  Each of the percentages being derived from a random sample
of 100 observations, we can compute confidence intervals under a
normality assumption; if we assume that the observations are
independent, then 95\% confidence intervals are narrower than one
twentieth of a percentage point for all the statistics computed.

The results show that up to 89\% of the translation lexicon entries
produced by SABLE on or above the 3rd likelihood plateau ``can be of
help in constructing a glossary.''  Up to 56\% can be considered
useful essentially as-is (the {\em V\/} category alone).  Including
all entries on the 2nd plateau or higher provides better coverage, but
reduces the fraction of useful entries to 81\%.  The fraction of
entries that are useful as-is remains roughly the same, at 55\%.  At
both recall levels, the extra Hansards-based filter had a detrimental
effect on precision.

Note that these figures are based on translation lexicons from which
many valid general usage entries have been filtered out (see
Section~\ref{sec:evaluation}).  We can compute SABLE's precision on
unfiltered translation lexicons for this corpus by assuming that
entries appearing in the Collins MRD are all correct.\footnote{Result:
88.4\% precision at 37.0\% recall or 93.7\% precision at 30.4\%
recall.}  However, these are not the real figures of interest here,
because we are mainly concerned in this study with the acquisition of
domain-specific translation lexicons.

\subsection{Recall}  

Following Melamed \shortcite{mel96b}, we adopt the following approach
to measuring recall: the upper bound is defined by the number of
different words in the bitext.  Thus, perfect recall implies at least
one entry containing each word in the corpus.  This is a much more
conservative metric than that used by Daille et
al. \shortcite{daille1994}, who report recall with respect to a
relatively small, manually constructed reference set.  Although we do
not expect to achieve perfect recall on this criterion after general
usage entries have been filtered out, the number is useful insofar as
it provides a sense of how recall for this corpus correlates with
precision.  We have no reason to expect this correlation to change
across domain-specific and general lexicon entries.  For the
unfiltered translation lexicons, recall on the \mbox{3rd} likelihood
plateau and above was 30.4\%.  When all entries on and above the
\mbox{2nd} plateau were considered, recall improved to 37.0\%.

\begin{figure}[htb]
\centerline{\psfig{figure=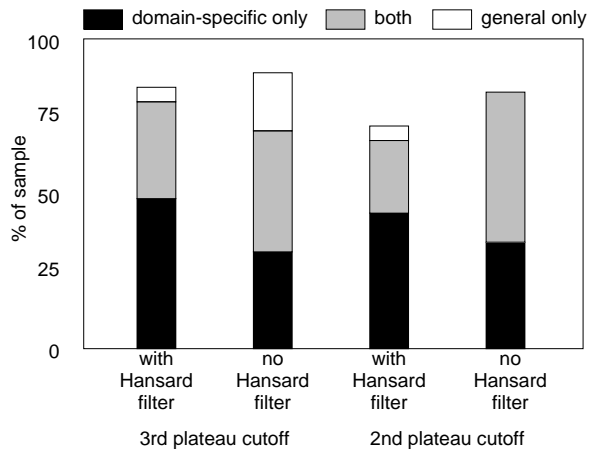,width=3in}}
\caption{{\em Summary of filtered translation lexicon domain-specificity
statistics.}}
\label{domain}
\end{figure}
\begin{table}[htb]
%\begin{small}
\begin{tabular}{|c|c||c|c|c|} \hline
& & \% & \%  & \\
Hansards & Plateau & Domain & General & \% \\
Filter? & Cutoff  & Specific & Usage & Both  \\
\hline\hline
Yes & 3rd & 
  82  & 37  & 35  \\
No & 3rd & 
  71  & 53  & 35 \\
Yes & 2nd & 
  66  & 27  & 22 \\
No & 2nd & 
  81  & 47  & 47 \\
\hline
\end{tabular}
%\end{small}
\caption{{\em Domain-specificity of filtered translation lexicon entries.}
    \label{tbl:domain}}
\end{table}

\subsection{Domain Specificity}

Figure~\ref{domain} demonstrates the effectiveness of the MRD- and
corpus-based filters, with details in Table~\ref{tbl:domain}.  If we
assume that translation pairs in the Collins MRD are not specific to
our chosen domain, then domain-specific translation lexicon entries
constituted only 49\% of SABLE's unfiltered output on or above the 2nd
plateau and 41\% on or above the 3rd plateau.  The MRD filter
increased this ratio to 81\% and 71\%, respectively.  As noted in
Section~\ref{precision}, the second filter, based on the Hansard
bitext, reduced the overall accuracy of the translation lexicons.  Its
effects on the proportion of domain-specific entries was mixed: an
11\% increase for the entries more likely to be correct, but a 15\%
decrease overall.  The corpus-based filter is certainly useful in the
absence of an MRD.  However, our results suggest that combining
filters does not always help, and more research is needed to
investigate optimal filter combination strategies.

\subsection{Consistency of Annotations}

In order to assess the consistency of annotation, we follow Carletta
\shortcite{carletta1996} in using Cohen's $\kappa$, a chance-corrected
measure of inter-rater agreement.  The $\kappa$ statistic was developed
to distinguish among levels of agreement such as ``almost perfect,
substantial, moderate, fair, slight, poor'' \cite{agresti1992}, and
Carletta suggests that as a rule of thumb in the behavioral sciences,
values of $\kappa$ greater than .8 indicate good replicability, with
values between .67 and .8 allowing tentative conclusions to be drawn.
For each such comparison, four values of $\kappa$ were computed:
\begin{itemize}
\item[$\kappa_1$:] agreement on the evaluation of
whether or not a pair should be immediately rejected or retained;
\item[$\kappa_2$:] agreement, for the retained pairs, on the type V, P, or I
assigned to the pair;
\item[$\kappa_3$:] agreement, for the retained pairs, on whether to
classify the pair as being useful for constructing a domain-specific glossary;
\item[$\kappa_4$:] agreement, for the retained pairs, on whether to
classify the pair as being useful for constructing a general usage glossary.
\end{itemize}
In each case, the computation of the agreement statistic took into
account those cases, if any, where the annotator could not arrive at a
decision for this case and opted simply to throw it out.  
\begin{table}
\begin{center}
\begin{tabular}{|l|llllll|}\hline
$\kappa$  & A1	& A2	& A3 	& A4	& A5	& A6 \\ \hline\hline
$\kappa_1$ & 0.70 & 0.44 & 0.59 & 0.82 & 0.90 & 0.82 \\
$\kappa_2$ & 0.62 & 0.67 & 0.72 & 0.74 & 0.55 & 0.73 \\
$\kappa_3$ & 0.28 & 0.19 & 0.50 & 0.00 & 0.00 & 0.56 \\
$\kappa_4$ & 0.67 & 0.69 & 0.68 & 0.74 & 0.61 & 0.81 \\
\hline
\end{tabular}
\end{center}
\caption{{\em Inter-annotator agreement.}
	\label{tbl:reliability}}
\end{table}
Resulting values for inter-rater reliability are shown in
Table~\ref{tbl:reliability}; the six annotators are identified
as A1, A2, $\ldots$ A6, and each value of $\kappa$ reflects the
comparison between that annotator and the group annotation.  

With the exception of $\kappa_3$, these values of $\kappa$ indicate that
the reliability of the judgments is generally reasonable, albeit not
entirely beyond debate.  The outlandish values for $\kappa_3$, despite
high rates of absolute agreement on that dimension of annotation, are
explained by the fact that the $\kappa$ statistic is known to be highly
problematic as a measure of inter-rater reliability when one of the
categories that can be chosen is overwhelmingly likely
\cite{grove1981,spitznagel1985}.  Intuitively this is not surprising:
we designed the experiment to yield a predominance of domain-specific
terms, by means of the MRD and Hansards filters.  Our having
succeeded, there is a very high probability that the ``Specific''
annotation will be selected by any two annotators, because it appears
so very frequently; as a result the actual agreement rate for that
annotation doesn't actually look all that different from what one
would get by chance, and so the $\kappa$ values are low.  The values of
$\kappa_3$ for annotators~4 and~5 emphasize quite clearly that $\kappa$
is measuring not the level of absolute agreement, but the
distinguishability of that level of agreement from chance.

\section{Conclusion}

In this paper, we have investigated the application of SABLE, a
turn-key translation lexicon construction system for non-technical
users, to the problem of identifying domain-specific word translations
given domain-specific corpora of limited size.  Evaluated on a very
small (400,000 word) corpus, the system shows real promise as a method
of processing small domain-specific corpora in order to propose
candidate single-word translations: once likely general usage terms
are automatically filtered out, the system obtains precision up to
89\% at levels of recall very conservatively estimated in the range of
30--40\% on domain-specific terms.

Of the proposed entries not immediately suitable for inclusion in a
translation lexicon, many represent part-of-speech divergences (of the
{\em protect/prot\'eg\'e\/} variety) and a smaller number incomplete
entries (of the {\em imm\'ediatement/right\/} variety) that would
nonetheless be helpful if used as the basis for a bilingual
concordance search --- for example, a search for French segments
containing {\em imm\'ediatement\/} in the vicinity of English segments
containing {\em right\/} would most likely yield up the obvious
correspondence between {\em imm\'ediatement\/} and {\em right away}.
Going beyond single-word correspondences, however, is a priority for
future work.

\section{Acknowledgments}

The authors wish to acknowledge the support of Sun Microsystems
Laboratories, particularly the assistance of Gary Adams, Cookie
Callahan, and Bob Kuhns, as well as useful input from Bonnie Dorr,
Ralph Grishman, Marti Hearst, Doug Oard, and three anonymous
reviewers. Melamed also acknowledges grants ARPA N00014-90-J-1863 and
ARPA N6600194C 6043.

%%%%%%%%%%%%%%%%%%%%%%%%%%%%%%%%%%%%%%%%%%%%%%%%%%%%%%%%%%%%%%%%
%%%
%%% For final submission, insert paper.bbl here instead
%%% of bibtex directives
%%%
%%%%%%%%%%%%%%%%%%%%%%%%%%%%%%%%%%%%%%%%%%%%%%%%%%%%%%%%%%%%%%%%

\bibliography{gisting,general,learning,distrib,nlstat,ibm_master,anlp94,multilingual}

\end{document}